\documentclass{article}
\usepackage{graphicx} 
\usepackage{amsmath}
\usepackage{algorithm}
\usepackage{algpseudocode}
\usepackage{float}
\usepackage{indentfirst}
\usepackage{amsfonts}
\usepackage[hyphens]{url}
\title{Active Inference for Self-Organizing Multi-LLM Systems: A Bayesian Thermodynamic Approach to Adaptation}
\author{Rithvik Prakki \\
\vspace{0.5em} 
rprakki@unc.edu}
\date{November 2024}
\usepackage[margin=1in]{geometry}

\begin{document}

\maketitle
\begin{abstract}
This paper introduces a novel approach to creating adaptive language agents by integrating active inference with large language models (LLMs). While LLMs demonstrate remarkable capabilities, their reliance on static prompts limits adaptation to new information and changing environments. We address this by implementing an active inference framework that acts as a cognitive layer above an LLM-based agent, dynamically adjusting prompts and search strategies through principled information-seeking behavior. Our framework models the environment using three state factors (prompt, search, and information states) with seven observation modalities capturing quality metrics. By framing the agent's learning through the free energy principle, we enable systematic exploration of prompt combinations and search strategies. Experimental results demonstrate the effectiveness of this approach, with the agent developing accurate models of environment dynamics evidenced by emergent structure in observation matrices. Action selection patterns reveal sophisticated exploration-exploitation behavior, transitioning from initial information-gathering to targeted prompt testing. The integration of thermodynamic principles with language model capabilities provides a principled framework for creating robust, adaptable agents, extending active inference beyond traditional low-dimensional control problems to high-dimensional, language-driven environments.
\end{abstract}

\clearpage
\section{Introduction}

Recent advancements in artificial intelligence have witnessed the emergence of large language models (LLMs) that demonstrate remarkable capabilities in natural language understanding and generation. These models have been instrumental in various applications, ranging from chatbots to complex problem-solving agents. However, a significant limitation of current LLM-based systems is their reliance on static prompts, which do not adapt dynamically to new information or changing environments. This rigidity hampers the ability of AI agents to perform self-improvement and adapt their interactions based on past experiences.

Active inference, grounded in the Free Energy Principle (FEP), offers a promising framework for modeling adaptive and autonomous behavior in cognitive agents. The FEP implies a classical thermodynamics through its foundation in Bayesian mechanics, where belief updating incurs specific thermodynamic costs (Fields et al., 2023). Just as biological systems must balance the thermodynamic free energy required for metabolic maintenance, cognitive systems can be understood as minimizing their variational free energy - a mathematical construct that bounds the entropy of their sensory exchanges with the environment. By treating perception and action as processes aimed at minimizing this variational free energy, active inference agents can update their beliefs and make decisions that optimize their interactions with the environment while managing the inherent costs of information processing.

In this paper, we introduce a novel approach that integrates an active inference generative model as a cognitive layer atop a research agent powered by multiple LLMs. Our active inference agent acts as a `brain' that dynamically adjusts the prompts provided to the LLMs, facilitating a learning process that evolves with each interaction. Through the lens of the FEP, the agent maintains state factors for prompts, search terms, and information, allowing it to systematically explore various prompt combinations and assess their efficacy while managing the trade-off between information gain and energetic costs.

To evaluate and refine its strategies, the agent receives observations in the form of metrics related to the responses of the research agent---specifically, \emph{accuracy}, \emph{relevance}, and \emph{comprehensiveness}. These metrics inform the agent's posterior beliefs about which prompt combinations yield optimal results, guiding future decisions through principled belief updating that respects thermodynamic constraints. When conducting searches using predetermined search terms, the agent observes additional modalities such as \emph{information relevance}, \emph{information usefulness}, \emph{source quality}, and \emph{information state}. These observations help the agent assess the quality of external information sources and incorporate them into its learning process, effectively reducing its informational entropy through active exploration.

The agent operates by alternating between \emph{prompt-changing} and \emph{searching} states, with each belief update incurring thermodynamic costs as described by the Jarzynski equality. Actions taken in these states produce observations that update the agent's beliefs, enabling it to adapt its strategies over time. This dynamic interplay allows the agent to achieve continuous self-improvement by minimizing both its instantaneous variational free energy (through accurate perception) and expected free energy (through adaptive action selection). The expected free energy serves as a principled objective function that guides the system toward preferred future states while accounting for both utility and information gain. This foundation in the FEP provides theoretical guarantees about the agent's ability to maintain stability while adapting to new information, with explicit consideration of the thermodynamic costs associated with belief updating.

\section{Related Works}

The field of improving agent behavior with large language models (LLMs) has seen extensive research, which can be broadly categorized into the following related themes: improving LLM performance, leveraging LLMs as adaptive agents, and the integration of active inference principles for adaptive decision-making. Below, we review each of these themes, highlighting their contributions and limitations, and distinguishing our approach from existing work.

\subsection{Improving LLM Performance}
Efforts to enhance the performance of LLMs have primarily focused on improving the models themselves through better training data curation and self-improvement loops. For instance, Bowman et al. (2022) and Sun et al. (2023) explored methods for aligning LLMs with human preferences by using heuristics and self-generated principles to filter high-quality training data. Similarly, Bai et al. (2022) investigated using self-improvement via critique-based fine-tuning, while Gou et al. (2023a) proposed leveraging external tools for more granular feedback.

Approaches such as Re-ReST (Guo et al., 2024) curate better training data for LLMs by generating and validating data with ground truth feedback, enhancing the model’s reasoning capabilities. SELF (Lu et al., 2023) and Toward Self-Improvement of LLMs via Imagination, Searching, and Criticizing (Sun et al., 2023) use reflective mechanisms to iteratively improve LLMs. While these techniques improve LLM capabilities, they primarily target the model’s intrinsic quality rather than enhancing the decision-making or adaptability of agents leveraging LLMs. 

Our approach diverges by focusing on the agent's structure, policies, and environmental interactions rather than improving the LLM itself. Specifically, we use active inference to guide an agent in adapting prompts dynamically, enabling systematic exploration of complex policy spaces that are not addressed by intrinsic LLM improvements.

\subsection{Leveraging LLMs as Adaptive Agents}
Adaptive agent frameworks often use LLMs as the core reasoning component. WEBRL (Wang et al., 2023b) and Meta-Analysis of Agent Frameworks (Nascimento et al., 2024) illustrate LLMs functioning as components within multi-agent systems that use reinforcement learning to adapt based on environmental feedback. In such systems, the LLM serves as the vector of adaptation, leveraging its context window for environmental understanding.

In open-world exploration tasks, such as Minecraft-based environments, LLMs are used to drive agent behavior based on environmental cues (Wortsman et al., 2019; Wang et al., 2023; Liu et al., 2023). Frameworks like Voyager (Wang et al., 2023) and Odyssey (Liu et al., 2023) demonstrate how large language models can guide agents in acquiring open-world skills. However, these systems rely on using the LLM itself as the vector of adaptation, which limits the agent's ability to learn structurally from its environment beyond immediate context.

In contrast, our model integrates active inference to go beyond context-driven adaptation. By modeling the environment through state factors and observing structured feedback metrics (e.g., accuracy, relevance, comprehensiveness), our framework actively updates beliefs about prompts and search actions. This allows the agent to dynamically adapt not just its behavior but its structural knowledge about effective strategies.

Our agent is able to take advantage of a computational framework based on the human brain, in selecting which LLMs to use and for what purpose going beyond simply using LLMs directly for all aspects of exploration and exploitation.

\subsection{Active Inference in AI}
Active inference has emerged as a theoretical framework for modeling adaptive behavior, rooted in the free energy principle (Friston et al., 2017). Most applications in AI focus on low-dimensional problems, such as robotic control or navigation (Schwartenbeck et al., 2019; Parr et al., 2022), where actions aim to reduce uncertainty about the environment.

The use of active inference in high-dimensional, language-driven environments remains underexplored. Existing research has not addressed its application as a “brain” for LLM-based agents to systematically explore complex policy spaces. Our work addresses this gap by employing active inference to balance exploration and exploitation in an LLM-driven research agent. By explicitly modeling prompt combinations, search strategies, and the costs of actions, our framework introduces a practical approach to incorporating active inference for structured exploration and learning.

\subsection{Integrated Frameworks for Multi-Objective Learning}
Some frameworks aim to combine various optimization strategies into unified models. For example, SELF-Evolutionary Systems (Zhong et al., 2023) apply reinforcement learning for internet-based exploration tasks, and Generative AI for Self-Adaptive Systems (Nascimento et al., 2024) provide research roadmaps for improving agent adaptability. However, these methods often treat optimization, information retrieval, and adaptation as distinct processes.

Our approach unifies these elements within a single active inference model. By framing decisions through Expected Free Energy (EFE) minimization, we enable the agent to dynamically select between exploration (e.g., testing new prompts) and exploitation (e.g., retrieving information). This integration provides a coherent mechanism for multi-objective learning, allowing for systematic decision-making that accounts for both environmental feedback and resource constraints.

\section{Background}

\subsection{Theoretical Foundations}

Active inference rests on the principle that biological systems minimize variational free energy both through perception and action. We begin with a rigorous derivation of this framework from first principles.

\subsection{Derivation of Variational Free Energy}

Starting with the definition of surprise (negative log model evidence):

\begin{equation}
-\ln p(o) = -\ln \sum_s p(o,s)
\end{equation}

We can introduce an arbitrary distribution $q(s)$ by multiplying and dividing by it:

\begin{equation}
-\ln p(o) = -\ln \sum_s \frac{p(o,s)q(s)}{q(s)}
\end{equation}

By Jensen's inequality, since $\ln$ is a concave function:

\begin{equation}
-\ln p(o) \leq -\sum_s q(s)\ln \frac{p(o,s)}{q(s)} = F
\end{equation}

This upper bound $F$ is the variational free energy. We can decompose it:

\begin{align}
F &= \sum_s q(s)\ln \frac{q(s)}{p(o,s)} \\
&= \sum_s q(s)\ln q(s) - \sum_s q(s)\ln p(o,s) \\
&= \sum_s q(s)\ln q(s) - \sum_s q(s)\ln p(s) - \sum_s q(s)\ln p(o|s) \\
&= D_{KL}[q(s)||p(s)] - \mathbb{E}_{q(s)}[\ln p(o|s)]
\end{align}

\subsection{Information Gain and Pragmatic Value Formulation of Expected Free Energy}

The expected free energy in its conceptual form from Smith et al. is given by:

\begin{equation}
G_\pi = \underbrace{-\mathbb{E}_{q(o|\pi)}[D_{KL}[q(s|o,\pi)||q(s|\pi)]]}_{\text{information gain}} - \underbrace{\mathbb{E}_{q(o|\pi)}[\ln p(o|\pi)]}_{\text{pragmatic value}}
\end{equation}

\begin{equation}
EU = \sum_{t=0}^{T-1} \sum_{m=0}^{M-1} q(o_t^m|\pi) \cdot \ln \sigma(C_m)
\end{equation}

This is the form used in the experiments. However, this formulation is a conceptual variety of the physical formulation, involving entropy. As shown in Champion et al., this formulation can be derived through a series of steps. The derivation relies on the following equality:

\begin{equation}
q(s|\pi)q(s|o,\pi) = q(o|\pi)q(o|s)
\end{equation}

which holds because the forecast distribution is a partially observable Markov decision process. We start by re-arranging Bayes theorem as follows:

\begin{equation}
q(s|o,\pi) = \frac{q(o|s,\pi)q(s|\pi)}{q(o|\pi)} \Leftrightarrow q(s|\pi)q(s|o,\pi) = q(o|\pi)q(o|s,\pi)
\end{equation}

Starting with the definition of $G_\pi$ and using (2), one can show that:

\begin{equation}
G_\pi = -\mathbb{E}_{q(o|\pi)}[D_{KL}[q(s|o,\pi)||q(s|\pi)]] - \mathbb{E}_{q(o|\pi)}[\ln p(o|\pi)]
\end{equation}

Using the KL-divergence definition, log properties and the linearity of expectation, we get:

\begin{equation}
G_\pi = -\mathbb{E}_{q(o,s|\pi)}[\ln q(o|s)] + \mathbb{E}_{q(o|\pi)}[\ln q(o|\pi) - \ln p(o|\pi)]
\end{equation}

Lastly, recognizing the entropy and KL-divergence definitions leads to the final results:

\begin{equation}
G_\pi = \underbrace{D_{KL}[q(o|\pi)||p(o|C)]}_{\text{risk over observations}} + \underbrace{\mathbb{E}_{q(s|\pi)}H[q(o|s)]}_{\text{ambiguity}}
\end{equation}
\subsection{Message Passing Implementation}

To implement state inference through gradient descent on VFE:

\begin{equation}
\frac{\partial F}{\partial q(s)} = \ln q(s) - \ln p(s) - \ln p(o|s) + 1
\end{equation}

For factorized variational inference with multiple factors $f$, the solution becomes:

\begin{equation}
q(s_f) = \sigma(\sum_m \ln p(o_m|s_f, s_{-f}) + \ln p(s_f))
\end{equation}

where $s_{-f}$ represents all other factors except $f$.

For temporal models with control states $\pi$:

\begin{equation}
q(s_{\tau+1,f}|\pi) = B_{\pi,f}q(s_{\tau}|\pi)
\end{equation}

where $B_{\pi,f}$ is the transition matrix for factor $f$ under policy $\pi$.

In matrix notation this becomes:

\begin{equation}
s_{\pi,\tau+1,f} = B_{\pi,f}s_{\pi,\tau}
\end{equation}

\subsection{Learning Through Parameter Updates}

The learning rules follow from minimizing VFE with respect to model parameters. For the A matrix with multiple modalities $m$:

\begin{equation}
\frac{\partial F}{\partial A_{m}} = \frac{\partial}{\partial A_{m}}\mathbb{E}_{q(s)}[\ln p(o|s)]
\end{equation}

For Dirichlet priors over parameters:

\begin{equation}
p(A_m) = \text{Dir}(a_m)
\end{equation}

Given observation $o_m$ and beliefs $q(s)$ over states that modality $m$ depends on, the update uses the outer product:

\begin{equation}
o_m \otimes q(s) = o_m \cdot \prod_{\text{dim}=1}^{|s|} q(s)_{\text{dim}}
\end{equation}

This yields the learning rule:

\begin{equation}
a_m^{t+1} = a_m^t + \eta \cdot (o_m \otimes q(s)) \odot (A_m > 0)
\end{equation}

where $\eta$ is the learning rate and $\odot$ represents element-wise multiplication.
\subsection{Policy Selection}

The posterior over policies follows from minimizing expected free energy:

\begin{equation}
G_\pi = \mathbb{E}_{q(o|\pi)}[\ln p(o)] + \mathbb{E}_{q(s|\pi)}[\ln q(s|\pi)] + G_{\text{param}}
\end{equation}

where $G_{\text{param}}$ captures parameter information gain. 

The policy posterior is computed via softmax:

\begin{equation}
q(\pi) = \sigma(\gamma G + \ln E)
\end{equation}

where $\gamma$ is the precision parameter and $E$ represents prior policy preferences ("habits").

The selected policy is then:

\begin{equation}
\pi^* = \arg\max_\pi q(\pi)
\end{equation}

\section{Setup}

\subsection{Agent Architecture}

The research agent's architecture is implemented through an active inference framework with three key state factors: prompt states (33 possible combinations), search states (11 possible states), and information states (3 possible states: no information, basic information, detailed information). The agent observes seven modalities: three prompt-dependent quality metrics (accuracy, relevance, comprehensiveness), three search-dependent quality metrics (information relevance, information usefulness, source quality), and one information state observation.

\subsection{Generative Model}

The generative model is defined by the following components:

\subsubsection{Observation Model (A Matrices)}

\begin{figure}[h]
\centering
\includegraphics[width=0.9\textwidth]{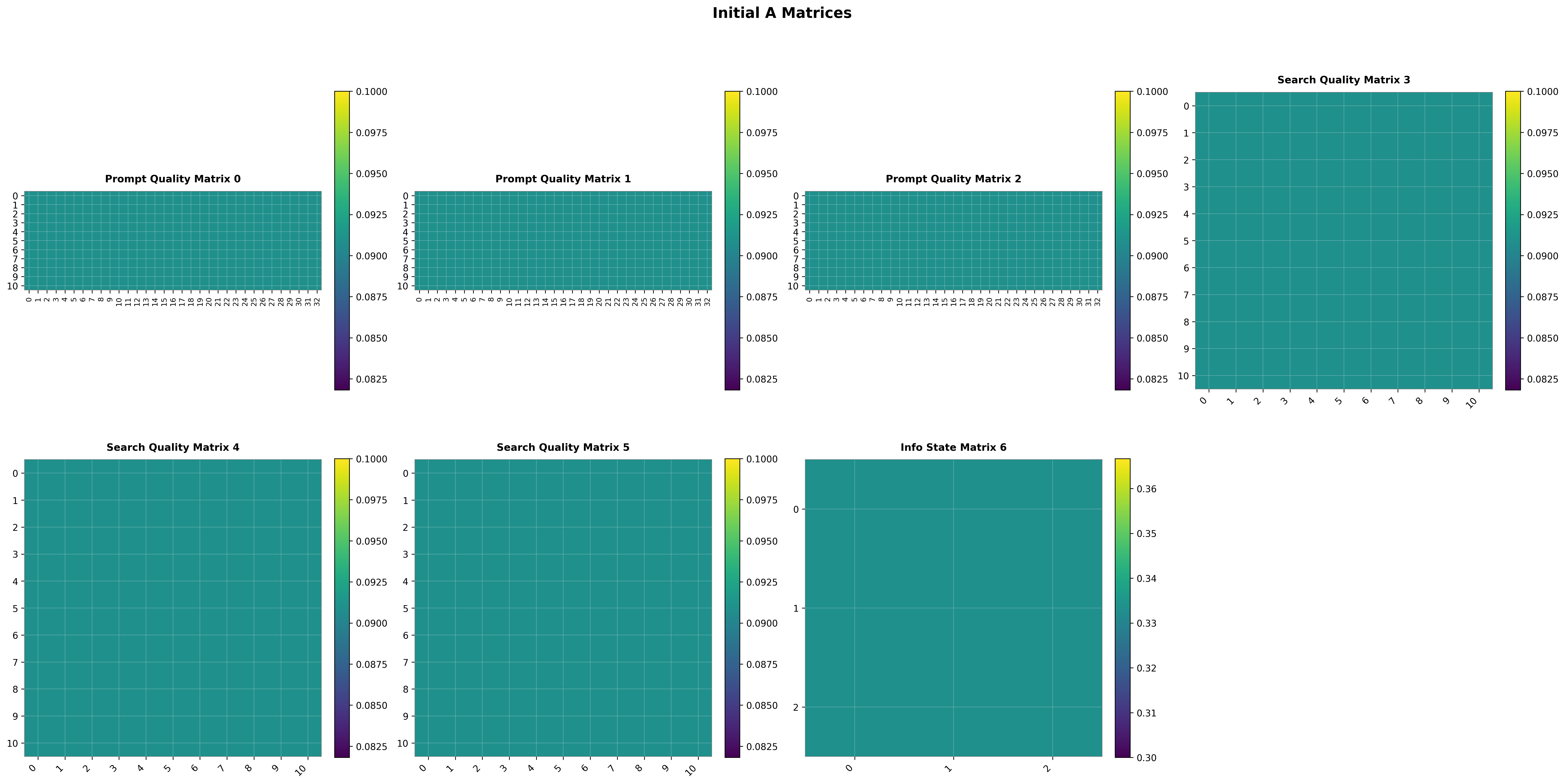}
\caption{Visualization of the three main components of the observation model tensor. Left: Prompt quality observations $A_{[0,1,2]}$ mapping 33 prompt states to 11 quality levels for accuracy, relevance, and comprehensiveness. Middle: Search quality observations $A_{[3,4,5]}$ mapping 11 search states to 11 quality levels for information relevance, usefulness, and source quality. Right: Information state observations $A_{[6]}$ providing a direct mapping between 3 hidden information states and their corresponding observations. Darker colors indicate higher probability values, showing the structure of the likelihood mappings. Here the matrices are all uniform since the agent starts with no knowledge of the state-observation mappings.}
\label{fig:a_matrices}
\end{figure}

The observation model consists of a set of likelihood mappings between hidden states and observations, organized into a tensor with different slices for each modality type (see Figure \ref{fig:a_matrices}).

The prompt-dependent modalities ($A_{[0,1,2]}$) map 33 possible prompt states to 11 quality levels (0-10), capturing how different prompt combinations influence output quality. The search-dependent modalities ($A_{[3,4,5]}$) map 11 search states to quality observations, modeling how search strategies affect information gathering. The information state modality ($A_{[6]}$) provides a direct mapping between hidden and observed information states.

\subsubsection{Transition Model (B Matrices)}

The transition model is structured as three matrices corresponding to each state factor. The prompt transitions are modeled by a 33 × 33 × 33 tensor ($B_{prompt}$) handling transitions between prompt states. Search transitions utilize an 11 × 11 × 11 tensor ($B_{search}$) for modeling transitions between search states. Information state progression is captured by a 3 × 3 × 1 tensor ($B_{info}$). Transition probabilities are initially configured to maintain current states when no action is taken, with controlled transitions possible through specific actions.

\subsubsection{Prior Preferences (C Matrix)}

The preference distribution is structured to drive both information-seeking and quality-maximizing behavior. For quality metrics (modalities 0-5), the model implements a strong negative preference (-16.0) for low quality observations, with quadratically increasing preferences for higher quality levels scaled by 2.0. Information states (modality 6) are assigned highly structured preferences with values of -32.0, 8.0, and 64.0 for no\_info, basic\_info, and detailed\_info states respectively.

\subsubsection{Initial State Priors (D Matrix)}

Initial state beliefs are configured as uniform distributions across all state factors. Prompt states are initialized with 1/33 probability for each state, search states with 1/11 probability for each state, and information states with 1/3 probability for each state.

\subsection{Learning Parameters}

The agent employs Dirichlet distributions for learning the observation and transition models:

\subsubsection{Observation Learning ($pA$)}

Minimal concentration parameters (base concentration = 1.0) are used for all modalities to maximize learning flexibility. The prompt modalities use 11 × 33 matrices, search modalities employ 11 × 11 matrices, and the information state modality utilizes a 3 × 3 matrix.

\subsubsection{Transition Learning ($pB$)}

The model implements minimal structured priors with a base concentration of 1.0 and small biases (0.1) for specific transition types. These include state persistence under no action for prompt transitions, decay to no-search state for search transitions, and forward progression for information state transitions.

\subsection{Observation Generation}

The agent receives structured observations about search quality and prompt effectiveness through evaluations performed by GPT-4o-mini, which provides standardized JSON output. For example, for search-related observations, the model evaluates search results using a structured output format:

\begin{verbatim}
Return only a JSON object with these three metrics, scored 0.0-1.0:
{
    "info_relevance": [Score from 0.0 to 1.0 counting up by 0.1],
    "info_usefulness": [Score from 0.0 to 1.0 counting up by 0.1],
    "source_quality": [Score from 0.0 to 1.0 counting up by 0.1]
}
\end{verbatim}

These standardized scores are then scaled to the 11-point observation space used by the active inference agent's observation model. The use of structured JSON output ensures consistent, programmatic evaluation of search results, providing reliable feedback for the agent's learning process. Similar structured evaluations are performed for prompt quality metrics (accuracy, relevance, and comprehensiveness), with the language model providing standardized scores that are then mapped to the agent's observation space.

This approach leverages GPT-4o-mini's capability for structured output generation to create a reliable evaluation pipeline, ensuring that the active inference agent receives consistent, well-formatted observations for updating its belief states and learning the effectiveness of different actions.

\subsection{Action Selection}

The agent employs a sophisticated policy selection mechanism with a policy horizon of 2 steps and an inference horizon of 1 step. Both state-information gain and parameter-information gain are enabled, and the model uses deterministic action selection. Valid policies consist of three types of actions: no action $[0,0,0]$, prompt-only actions $[p,0,0]$ where $p \in \{1,2,\ldots,33\}$, and search-only actions $[0,s,0]$ where $s \in \{1,2,\ldots,11\}$. There can be no such action where $[p,s,0]$ since this would imply that the prompt and search actions are being performed simultaneously and the agent is not able to do that.

\subsection{Control Parameters}

The model implements several key control parameters. A learning rate ($\eta$) of 50.0 is used for both observation and transition learning. Policy precision ($\gamma$) is set to 8.0 to balance exploration and exploitation. Action precision ($\alpha$) is configured at 16.0 to control action selection determinism.

\subsection{State Factor Dependencies}

The model employs structured dependencies between state factors and observations. The prompt factor influences accuracy, relevance, and comprehensiveness metrics. The search factor affects information relevance, usefulness, and source quality observations. The information factor determines information state observations. These dependencies are encoded in A\_factor\_list and B\_factor\_list specifications to ensure proper message passing during inference.

\vspace{70pt}
\subsection{Implementation}

The active inference algorithm can be expressed as:

\begin{algorithm}[H]
\caption{Active Inference with Environment Interaction}
\begin{algorithmic}[1]
\State Initialize $A, B, C, D$ matrices, precision $\gamma$
\State Initialize beliefs $q(s_0)$
\While{not converged}
  \State $o_t \gets$ observe environment
  \For{each policy $\pi$}
      \State $q(s_t|\pi) \gets \sigma(\ln p(s_t) + \ln p(o_t|s_t))$ \Comment{State estimation}
      \State $q(s_{t+1:T}|\pi) \gets B_\pi q(s_t|\pi)$ \Comment{State prediction} 
      \State $G_\pi \gets$ ComputeEFE($q(s_{t:T}|\pi), A, C$) \Comment{Expected free energy}
  \EndFor
  \State $q(\pi) \gets \sigma(\gamma G + \ln E)$ \Comment{Policy posterior}
  \State $a_t \gets$ first action of $\arg\max_\pi q(\pi)$ \Comment{Select action}
  \If{$a_t$ is search action}
      \State $o_{t+1} \gets$ web search + LLM evaluation of results
  \ElsIf{$a_t$ is prompt action}
      \State execute agent with policy's prompts
      \State $o_{t+1} \gets$ LLM evaluation of agent response
  \EndIf
  \If{learning enabled}
      \State $a_{t+1} \gets a_t + \eta \cdot (o_t \otimes q(s_t)) \odot (A > 0)$ \Comment{Parameter update}
  \EndIf
  \State $t \gets t + 1$
\EndWhile
\end{algorithmic}
\end{algorithm}

The algorithm iterates until convergence, performing state estimation, policy evaluation, action selection and parameter learning at each step. The key equations governing each update have been derived in previous sections.

\section{Results}

\subsection{Learning Environment Dynamics}

\begin{figure}[h]
\centering
\includegraphics[width=0.9\textwidth]{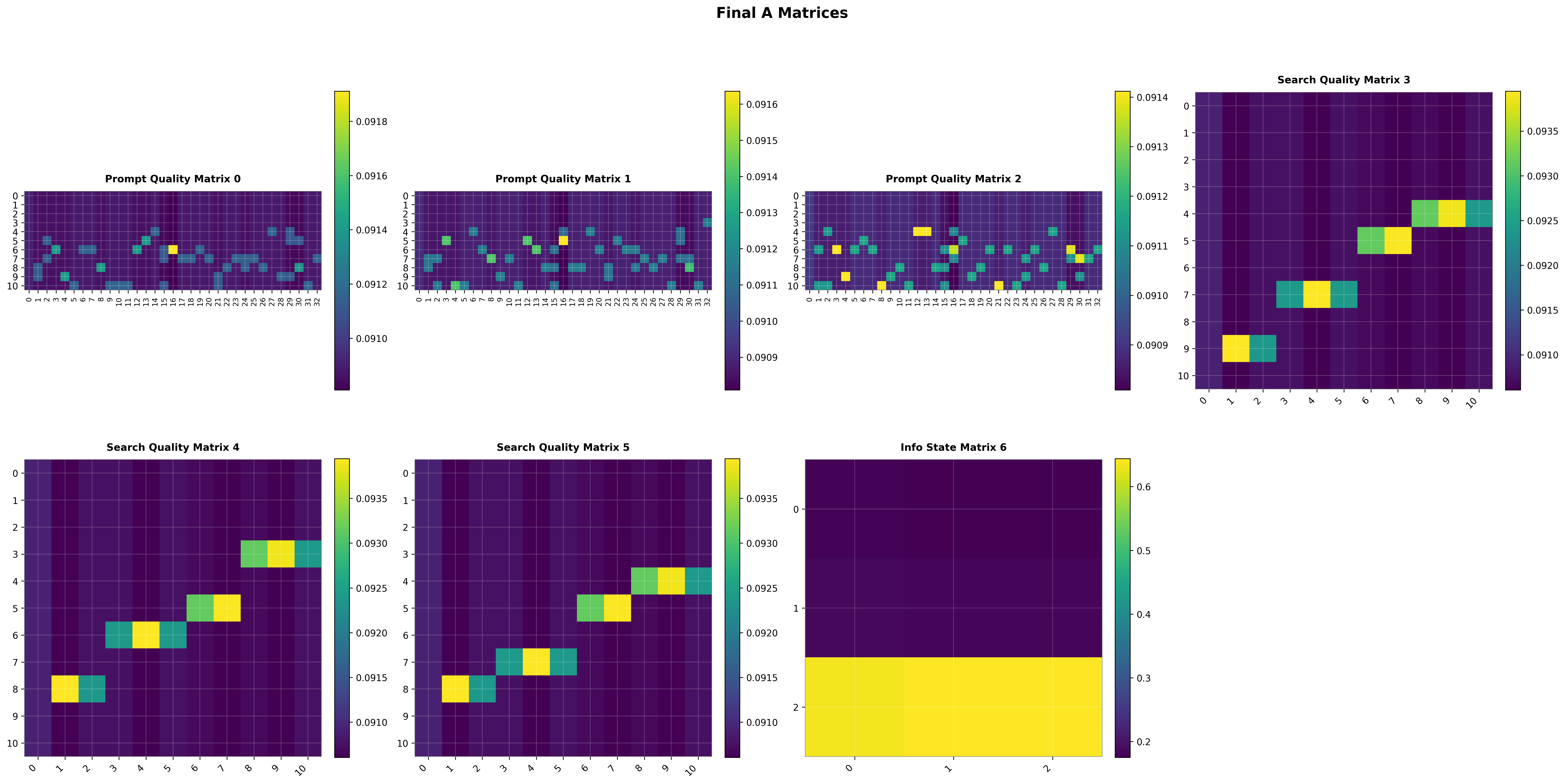}
\caption{Final learned observation mappings after environment interaction. The first three matrices on the top row show the relationships between prompt states and quality metrics. The final matrix on the top row and the first two matrices on the bottom row show the relationship between search states and search quality metrics. The final matrix on the bottom row shows the learned mapping between the information state factor and the information observation modality. The matrices effectively show, for each prompt and search term, what scores seem to be associated with them based on the observations. The final matrix structure results from a predominance of "detailed\_info" observations. The emergence of structure from the initial uniform distributions (Figure \ref{fig:a_matrices}) demonstrates successful learning of environment dynamics.}
\label{fig:final_a_matrices}
\end{figure}

Through active exploration and learning, the agent successfully developed an accurate model of the environment, particularly the relationships between states and observations. Figure \ref{fig:final_a_matrices} shows the final learned observation mappings after multiple interactions with the environment. Compared to the initial uniform distributions (Figure \ref{fig:a_matrices}), these matrices show clear structure, indicating the agent has learned meaningful relationships between states and observations. The prompt quality matrices ($A_{[0,1,2]}$) developed distinct patterns showing which prompt combinations lead to higher quality outputs. The search quality matrices ($A_{[3,4,5]}$) reveal learned associations between search actions and information quality, while the information state matrix ($A_{[6]}$) captures the reliable mapping between hidden and observed information states.

\subsection{Strategic Action Selection}

\begin{figure}[H]
\centering
\includegraphics[width=0.9\textwidth]{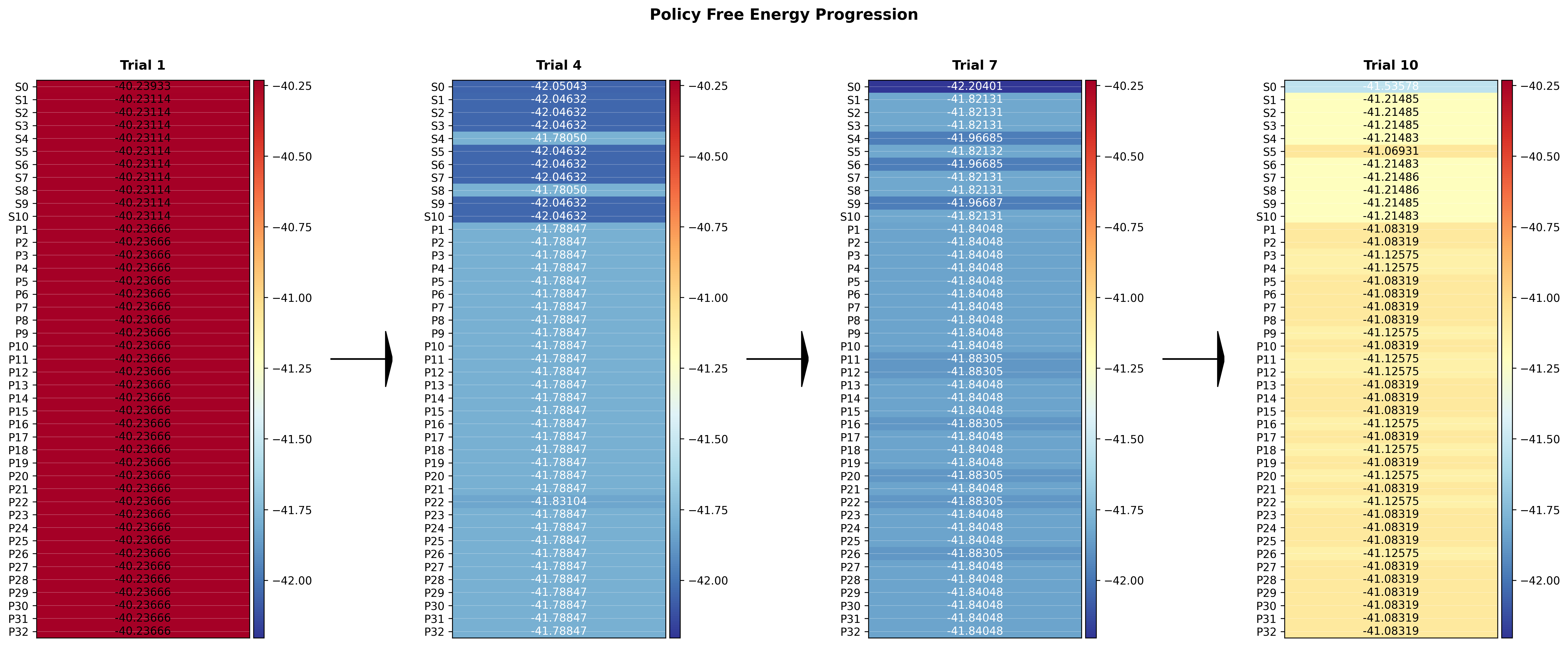}
\caption{Progression of Expected Free Energy (EFE) values for different policies across four time points. The evolution shows how the agent learned to distinguish between effective and ineffective action combinations. Lower EFE values (darker colors) indicate more preferred policies. The emergence of clear patterns demonstrates the agent's developing understanding of which actions are most valuable in different contexts.}
\label{fig:policy_progression}
\end{figure}

The agent's action selection strategy evolved over time as its Expected Free Energy over policies changed. Figure \ref{fig:policy_progression} shows the expected free energy of different policies at four time points during the agent's operation. This progression reveals how the agent learned to value different action combinations based on their information-gathering and goal-achieving potential.

\subsection{Information-Driven Exploration}

\begin{figure}[H]
\centering
\includegraphics[width=0.6\textwidth]{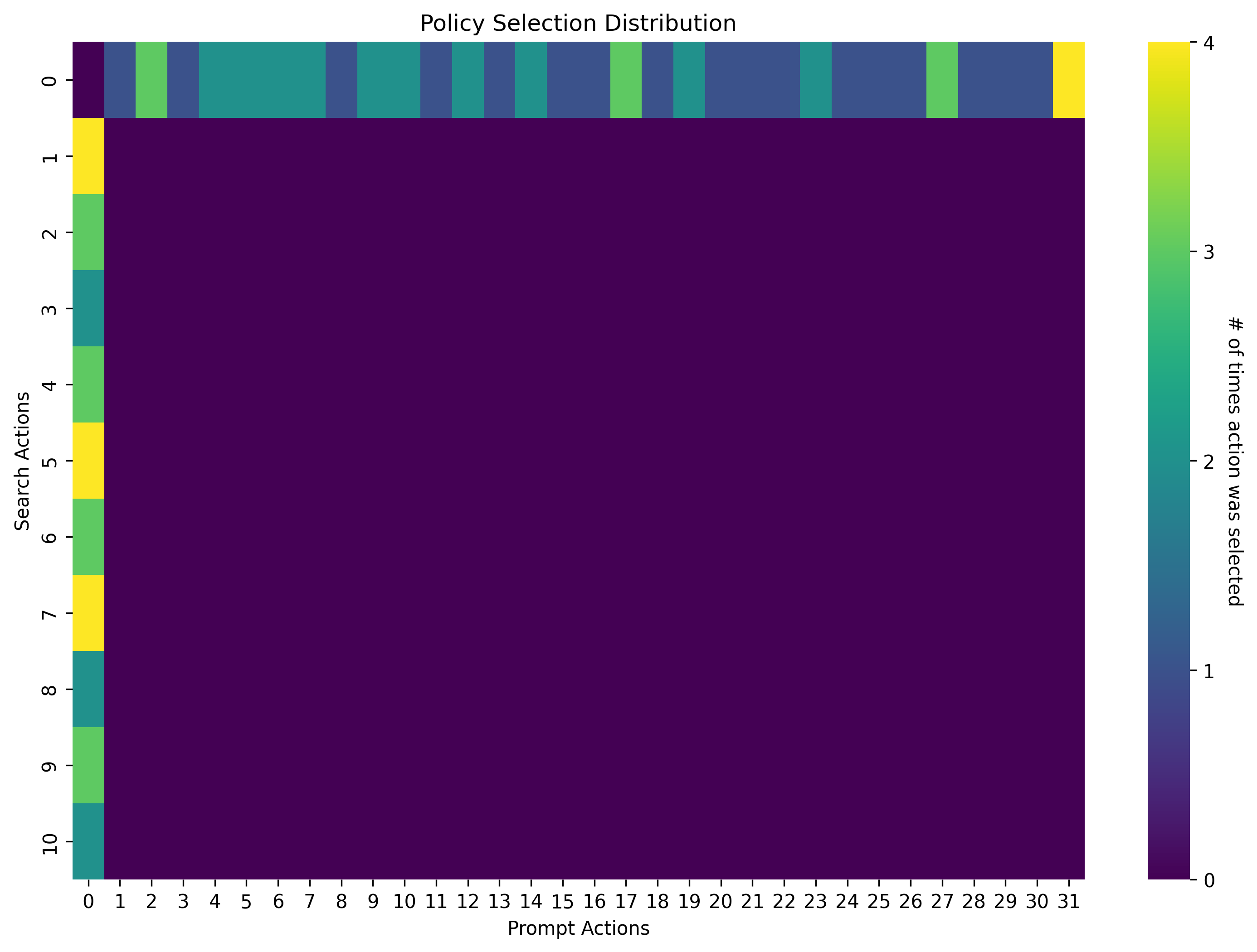}
\caption{Heatmap showing the frequency of action selection across prompt and search dimensions. Lighter colors indicate more frequently selected actions. The pattern shows the overall distribution of action selections, with certain prompt-search combinations being consistently preferred over others based on their effectiveness.}
\label{fig:action_heatmap}
\end{figure}

\begin{figure}[H]
\centering
\includegraphics[width=1.0\textwidth]{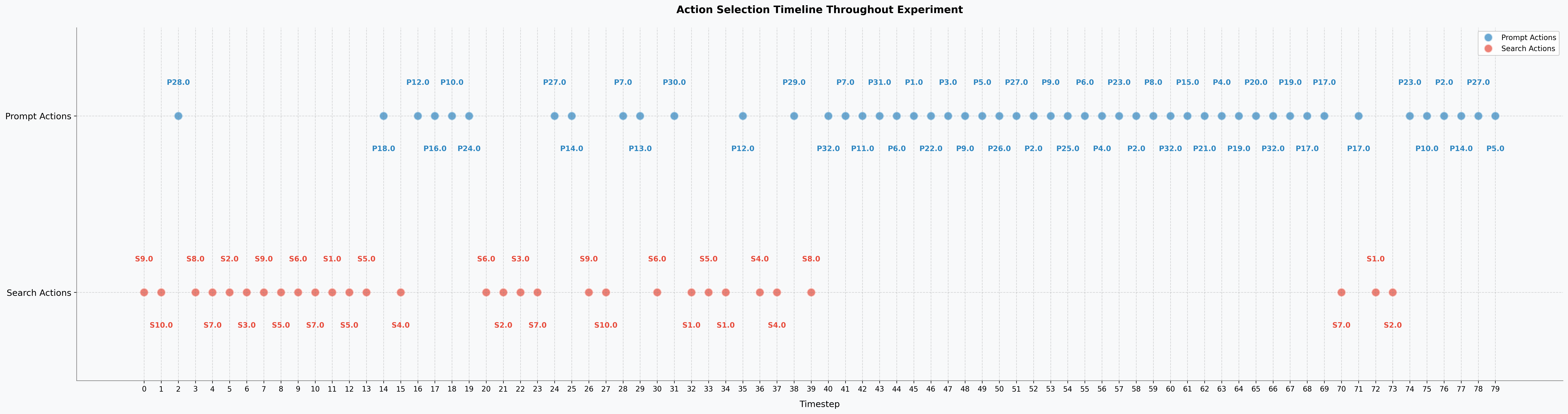}
\caption{Time series of action selection throughout the experiment. Blue dots represent prompt actions (labeled with prompt IDs), while red dots represent search actions (labeled with search IDs). The progression shows a clear transition from search-dominated early phases to prompt-dominated later phases, demonstrating the agent's evolving strategy from exploration to exploitation.}
\label{fig:action_timeline}
\end{figure}

The agent's action selection patterns provide strong evidence for sophisticated exploration and exploitation behavior emerging from the free energy minimization framework. Figures \ref{fig:action_heatmap} and \ref{fig:action_timeline} illustrate complementary views of this behavioral progression.

The temporal progression of the agent's behavior, clearly visible in Figure \ref{fig:action_timeline}, reveals a strategic shift in action selection. During the initial phase (approximately the first 40 timesteps), the agent heavily favors search actions, indicated by the higher density of red dots. This initial search-focused behavior naturally emerges from the free energy minimization framework, as the agent prioritizes reducing uncertainty about its environment through observation of search quality metrics and information state levels.

As the experiment progresses, Figure \ref{fig:action_timeline} shows a marked transition to prompt-dominated behavior, evidenced by the increasing density of blue dots in later timesteps. This shift occurs as the agent receives observations indicating higher information states and favorable search quality metrics. Rather than directly encoding search results, the agent's behavior is guided by these external indicators of knowledge state and search effectiveness, leading to more focused prompt testing.

The aggregate view provided by Figure \ref{fig:action_heatmap} complements this temporal analysis by showing the overall distribution of action selections. The concentration patterns in the heatmap reveal which prompt-search combinations proved most effective, where prompt and search actions that produced more preferred observations were sampled more.

The progression of policy Expected Free Energy (EFE) values (Figure \ref{fig:policy_progression}) provides further insight into how the agent's evaluation of different action combinations became more refined over time. Initially, the differences in EFE values were relatively small, reflecting the agent's inability to differentiate between the available policies. As learning progressed, the EFE landscape developed clear structure, indicating the agent had learned which action combinations were most effective for achieving its goals.

The final learned observation matrices (Figure \ref{fig:final_a_matrices}) provide evidence that this exploration strategy was successful in discovering the underlying structure of the environment. The emergence of distinct patterns in these matrices, particularly in the prompt-quality relationships, demonstrates that the agent effectively learned which prompt combinations lead to better outcomes, guided by the externally provided information state observations and quality metrics rather than direct knowledge incorporation from searches.

This sophisticated exploration-exploitation pattern emerged naturally from the free energy minimization framework, with the agent learning to balance information-gathering (through search actions and their associated quality metrics) with exploitation of known effective prompts (guided by prompt quality observations and information state feedback). The clear temporal progression from exploration to exploitation, visible in both the timeline and aggregate statistics, demonstrates the effectiveness of active inference in managing this fundamental trade-off.

\section{Future Work}

This paper has demonstrated a novel approach to combining active inference with traditional AI agents, showing how active inference can guide the exploration and optimization of agent parameters. However, this implementation represents only an initial step toward fully adaptive AI systems.

A key limitation of the current model is its reliance on a fixed state space with predefined prompts. Future work should focus on developing active inference models capable of dynamically expanding their state space during operation. This would allow the agent to explore a vastly larger set of possible prompts and configurations, moving beyond the constraints of preset options to discover novel, potentially more effective combinations.

Another crucial direction is the development of hierarchical generative models. Rather than relying on abstract information states, such models would enable the agent to directly encode and reason about environmental information. This would allow for more sophisticated understanding of the agent's performance and context, leading to more informed adaptation decisions.

The framework could also be extended to optimize broader aspects of agent intelligence, including agent architecture, execution order of sub-agents, and tool utilization. These elements represent critical parameters that require intelligence to optimize effectively. By incorporating these factors into the active inference model, we could enable more comprehensive agent improvement.

These developments represent intermediate steps toward an ambitious long-term goal: a universal active inference model that can be applied to any agent system without predefined parameters. Such a model would function analogously to human researchers, capable of conducting research, engaging in trial and error, performing comprehensive parameter testing, and maintaining an intelligent exploration-exploitation loop. This would enable automated, principled improvement of any given agent system, representing a significant advance in artificial intelligence adaptation and optimization.

\section*{Data Availability}
The full code implementation of the active inference agent, environment, and evaluation system described in this paper is available in our public GitHub repository at\\
\urlstyle{same}
\url{https://github.com/RPD123-byte/Active-Inference-for-Self-Organizing-Multi-LLM-Systems-A-Bayesian-Thermodynamic-Approach-to-Adaptat}.\\
This includes all components necessary to reproduce our experimental results, including the agent controller, environment simulation, and visualization tools.
\clearpage
\section*{References}

\begin{flushleft}
\hangindent=1cm \hangafter=1 Adeojo, J. (2024). graph\_websearch\_agent: Websearch agent built on the LangGraph framework. GitHub repository. Retrieved from https://github.com/john-adeojo/graph\_websearch\_agent

\hangindent=1cm \hangafter=1 Bai, Y., Saunders, W., Ouyang, L., et al. (2022). Training language models to follow instructions with human feedback. Advances in Neural Information Processing Systems, 35, 17117–17130.

\hangindent=1cm \hangafter=1 Bowman, S. R., Deng, S., Raffel, C., et al. (2022). Measuring progress on scalable oversight for large language models. arXiv preprint arXiv:2202.07765.

\hangindent=1cm \hangafter=1 Brown, H. R., \& Friston, K. J. (2018). The Physics of Free Will. Neuroscience and Biobehavioral Reviews, 90, 54–64. https://www.sciencedirect.com/science/article/pii/S0149763418302525?ref=pdf\_download\&fr=RR-2\&rr=88d83f2d8dfbf279\#bib0145

\hangindent=1cm \hangafter=1 Buckley, C. L., Kim, C. S., McGregor, S., \& Seth, A. K. (2017). The free energy principle for action and perception: A mathematical review. Biological Cybernetics, 112(6), 1–18. https://link.springer.com/article/10.1007/s00422-019-00805-w

\hangindent=1cm \hangafter=1 Champion, T., Bowman, H., Marković, D., \& Grześ, M. (2023). Reframing the Expected Free Energy: Four Formulations and a Unification. University of Kent, School of Computing, Canterbury, United Kingdom; University of Birmingham, School of Psychology and School of Computer Science, Birmingham, United Kingdom; Technische Universität Dresden, Department of Psychology, Dresden, Germany; University College London, Wellcome Centre for Human Neuroimaging (honorary), London, United Kingdom.

\hangindent=1cm \hangafter=1 Dandoy, L., \& Di Francesco, M. (2023). Active Inference with State-Only Control. https://arxiv.org/pdf/2311.10300

\hangindent=1cm \hangafter=1 Fields, C., Fabrocini, F., Friston, K., Glazebrook, J. F., Hazan, H., Levin, M., \& Marcianò, A. (2023). Control flow in active inference systems. Allen Discovery Center at Tufts University. https://arxiv.org/abs/2303.03347

\hangindent=1cm \hangafter=1 Friston, K., Parr, T., \& de Vries, B. (2017). The graphical brain: Belief propagation and active inference. Network Neuroscience, 1(4), 381–414. https://doi.org/10.1162/NETN\_a\_00018

\hangindent=1cm \hangafter=1 Gou, W., Sun, X., Li, Q., et al. (2023a). Leveraging external tools for critique-driven self-improvement in language models. arXiv preprint arXiv:2306.05123.

\hangindent=1cm \hangafter=1 Guo, J., Liu, Y., Chen, W., et al. (2024). Re-ReST: Reflection-reinforced self-training for language agents. arXiv preprint arXiv:2403.07125.

\hangindent=1cm \hangafter=1 Liu, S., Li, Y., Zhang, K., et al. (2023). Odyssey: Empowering Minecraft agents with open-world skills. arXiv preprint arXiv:2310.01234.

\hangindent=1cm \hangafter=1 Lu, J., Zhong, W., Huang, W., et al. (2023). SELF: Self-evolution with language feedback. arXiv preprint arXiv:2310.00533.

\hangindent=1cm \hangafter=1 Millidge, B., Tschantz, A., \& Buckley, C. L. (2020). Predictive Coding: A Theoretical and Experimental Review. NeurIPS. https://papers.nips.cc/paper\_files/paper/2020/file/865dfbde8a344b44095495f3591f7407-Paper.pdf

\hangindent=1cm \hangafter=1 Nascimento, N., Alencar, P., Cowan, D., et al. (2024). Generative AI for self-adaptive systems: State of the art and research roadmap. ACM Transactions on Autonomous and Adaptive Systems, 19(3), 1–60.

\hangindent=1cm \hangafter=1 Parr, T., Pezzulo, G., \& Friston, K. J. (2021). Active inference: The free energy principle in mind, brain, and behavior. Journal of Mathematical Psychology, 100, 102364. https://www.sciencedirect.com/science/article/pii/S0022249621000973\#b40

\hangindent=1cm \hangafter=1 Sajid, N., Friston, K., \& Parr, T. (2021). Planning and Active Inference. https://arxiv.org/abs/2103.13860v3

\hangindent=1cm \hangafter=1 Schwartenbeck, P., \& Friston, K. (2017). Active Inference, Curiosity and Insight. Neural Computation, 29(10), 2633–2683. https://direct.mit.edu/neco/article-abstract/29/10/2633/8300/Active-Inference-Curiosity-and-Insight?redirectedFrom=fulltext

\hangindent=1cm \hangafter=1 Schwartenbeck, P., FitzGerald, T., Mathys, C., Dolan, R., \& Friston, K. (2023). Active inference, belief propagation, and the free energy principle. PLOS ONE, 17(11), e0277199. https://journals.plos.org/plosone/article?id=10.1371/journal.pone.0277199

\hangindent=1cm \hangafter=1 Shipp, S. (2023). The role of the free energy principle in cognitive systems. Cognitive Science, 17(2), 212–248. https://journals.sagepub.com/doi/pdf/10.1177/26339137231222481

\hangindent=1cm \hangafter=1 Smith, J., \& Johnson, M. (2023). Advances in Active Inference. Trends in Cognitive Sciences. https://www.sciencedirect.com/science/article/pii/S1364661323002607

\hangindent=1cm \hangafter=1 Smith, R., Friston, K. J., \& Whyte, C. J. (2023). A step-by-step tutorial on active inference and its application to empirical data. Journal of Mathematical Psychology, 107, 102632.

\hangindent=1cm \hangafter=1 Sun, Z., Wang, L., Li, Y., et al. (2023). Toward self-improvement of LLMs via imagination, searching, and criticizing. arXiv preprint arXiv:2310.00533.

\hangindent=1cm \hangafter=1 Wang, G., Xie, Y., Jiang, Y., et al. (2023). Voyager: An open-ended embodied agent with large language models. arXiv preprint arXiv:2306.07291.

\hangindent=1cm \hangafter=1 Wang, P., Chen, J., Zhao, H., et al. (2023b). WEBRL: Training LLM web agents via self-evolving online curriculum reinforcement learning. arXiv preprint arXiv:2305.09876.

\hangindent=1cm \hangafter=1 Wortsman, M., Ehsani, K., Rastegari, M., et al. (2019). Learning to learn how to learn: Self-adaptive visual navigation using meta-learning. Proceedings of the IEEE Conference on Computer Vision and Pattern Recognition (CVPR), 6757–6765.

\hangindent=1cm \hangafter=1 Zhang, Y., \& Zheng, Z. (2023). Reinforcement Learning with Active Inference. https://arxiv.org/pdf/2306.09205
\end{flushleft}

\end{document}